\documentclass[sigconf]{acmart}

\usepackage{balance}

\def\BibTeX{{\rm B\kern-.05em{\sc i\kern-.025em b}\kern-.08emT\kern-.1667em\lower.7ex\hbox{E}\kern-.125emX}}
    
\copyrightyear{2019} 
\acmYear{2019} 
\acmConference[MM '19]{Proceedings of the 27th ACM International Conference on Multimedia}{October 21--25, 2019}{Nice, France}
\acmBooktitle{Proceedings of the 27th ACM International Conference on Multimedia (MM '19), October 21--25, 2019, Nice, France}
\acmPrice{15.00}
\acmDOI{10.1145/3343031.3350911}
\acmISBN{978-1-4503-6889-6/19/10}

\usepackage{subfig,multirow,makecell}
\usepackage{epsfig}
\usepackage{graphicx}
\usepackage{amsmath}
\usepackage{amssymb}
\usepackage{xspace}

\makeatletter
\DeclareRobustCommand\onedot{\futurelet\@let@token\@onedot}
\def\@onedot{\ifx\@let@token.\else.\null\fi\xspace}

\def\ie{\emph{i.e}\onedot} 

\def\etc{\textit{etc.}} 
 
\def\etal{\emph{et al}\onedot}
\makeatother

\newcommand{\app}{\raise.17ex\hbox{$\scriptstyle\sim$}}
\newcommand{\Caption}[1]{\caption{\small #1}\vspace{-2mm}}

\newcommand{\tss}[1]{\textsuperscript{#1}}

\newlength\savewidth\newcommand\shline{\noalign{\global\savewidth\arrayrulewidth
		\global\arrayrulewidth 1pt}\hline\noalign{\global\arrayrulewidth\savewidth}}

\begin{document}

\fancyhead{}

\title[Learning Sem-dist map with SLN for Amodal Instance Segmentation]{Learning Semantics-aware Distance Map with Semantics Layering Network for Amodal Instance Segmentation}

\author{Ziheng Zhang}
\authornote{indicates equal contributions.}
\email{zhangzh@shanghaitech.edu.cn}
\orcid{0000-0002-4496-1861}
\affiliation{%
	\institution{ShanghaiTech University}
	\city{Shanghai}
	\country{China}
}

\author{Anpei Chen}
\authornotemark[1]
\email{chenap@shanghaitech.edu.cn}
\affiliation{%
	\institution{ShanghaiTech University}
	\city{Shanghai}
	\country{China}
}

\author{Ling Xie}
\email{xieling@shanghaitech.edu.cn}
\affiliation{%
	\institution{ShanghaiTech University}
	\city{Shanghai}
	\country{China}
}

\author{Jingyi Yu}
\email{yujingyi@shanghaitech.edu.cn}
\affiliation{%
	\institution{ShanghaiTech University}
	\city{Shanghai}
	\country{China}
}

\author{Shenghua Gao}
\authornote{indicates the corresponding author.}
\email{gaoshh@shanghaitech.edu.cn}
\affiliation{%
	\institution{ShanghaiTech University}
	\city{Shanghai}
	\country{China}
}

%
\renewcommand{\shortauthors}{Zhang and Chen, \etal}

%
\begin{abstract}
In this work, we demonstrate yet another approach to tackle the amodal segmentation problem. Specifically, we first introduce a new representation, namely a semantics-aware distance map (sem-dist map), to serve as our target for amodal segmentation instead of the commonly used masks and heatmaps. The sem-dist map is a kind of level-set representation, of which the different regions of an object are placed into different levels on the map according to their visibility. It is a natural extension of masks and heatmaps, where modal, amodal segmentation, as well as depth order information, are all well-described. Then we also introduce a novel convolutional neural network (CNN) architecture, which we refer to as semantic layering network, to estimate sem-dist maps layer by layer, from the global-level to the instance-level, for all objects in an image. Extensive experiments on the COCOA and D2SA datasets have demonstrated that our framework can predict amodal segmentation, occlusion, and depth order with state-of-the-art performance.
\end{abstract}

%
%


%

%

%
\maketitle

\section{Introduction}
\begin{figure}[t]
\centering
\includegraphics[width=\linewidth]{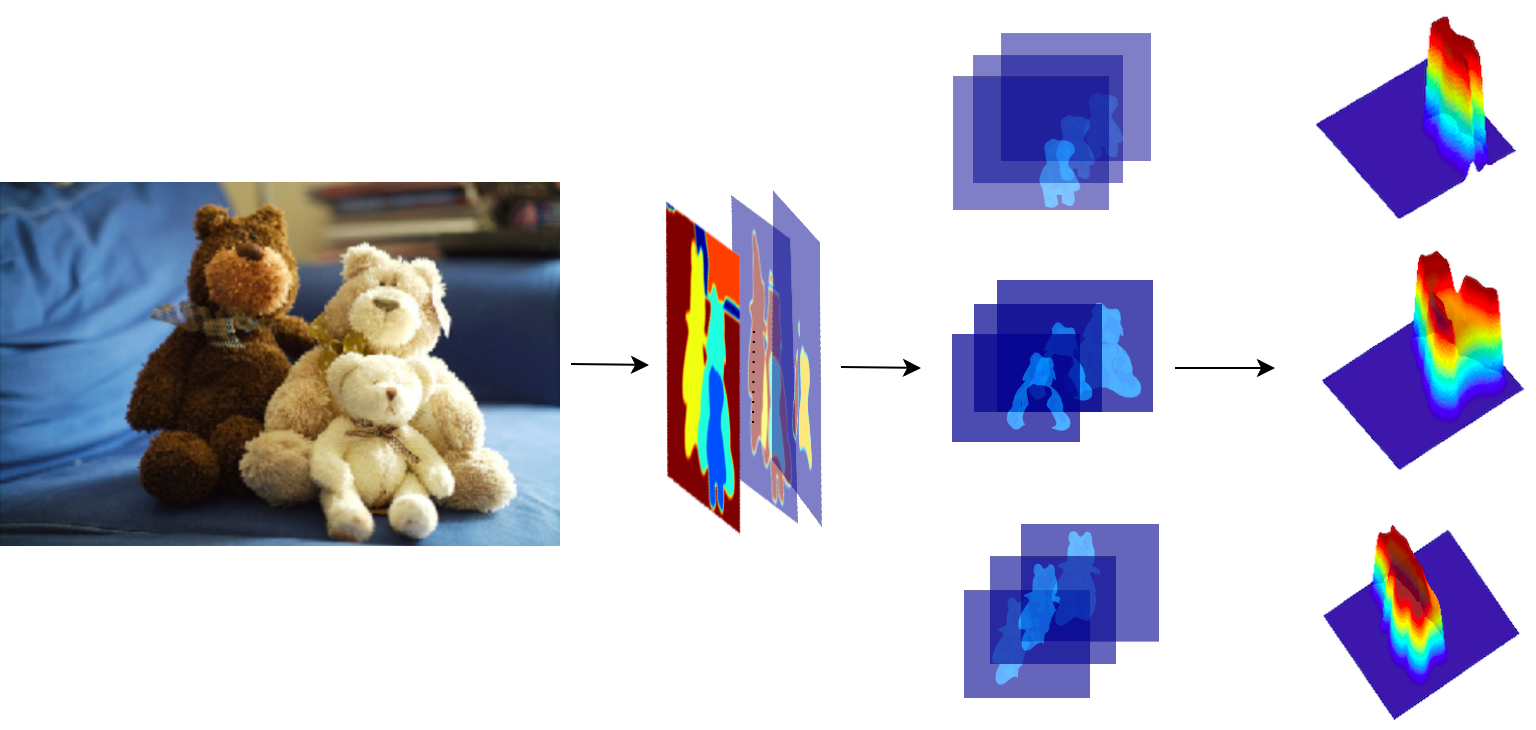}
\caption{The pipeline of our framework. The pictures from left to right are: 1) an input image; 2) the global layering map, which describes the number of overlapped objects in each pixel of the input image; 3) the instance layering maps, which contain the pixel-wise depth layer indices for the corresponding object instance in the input image; 4) the sim-dist maps, which indicate the modal segmentation, the amodal segmentation and the relative depth order of all object instances in the input image. An amodal segmentation of an object is inferred layer by layer, from global layering maps to instance layering maps and the final sem-dist maps in SLN (Semantics Layering Network). (best viewed in color)}
\label{fig:pipeline} 
\end{figure}
The recent years have witnessed great progress in visual understanding from image classification~\cite{krizhevsky2012imagenet,simonyan2014very,he2016deep} and detection~\cite{ren2015faster,liu2016ssd,lin2017feature}  to segmentation~\cite{chen2014semantic,long2015fully,chen2018deeplab}. It appears that the performance of machine vision systems is stepping closer and closer to that of humans in terms of accuracy. Despite that, human vision has the strong ability to see beyond the visible, i.e., to perceive whole semantic concepts with only partial visibility. This ability, also known as amodal perception~\cite{zhu2017semantic}, had hardly been exploited to develop machine vision systems with similar capability until most recently, when researchers started making such attempts by modifying state-of-the-art segmentation models and training them with synthetic~\cite{li2016amodal,zhu2017semantic,follmann2019learning} and/or human-annotated datasets~\cite{zhu2017semantic,follmann2019learning}.

Amodal perception implicitly requires vision systems to possess three critical abilities. The first is to recognize an object even if it is partially occluded by others. The second is to infer the most probable appearance of the invisible parts of an object given only the visible parts. And the third is to be aware of the relationship between overlapping objects. In other words, an amodal perception system should always keep a \textit{consistent} whole picture of an object given arbitrary occlusion patterns; and it should understand the depth order between objects. Indeed, every single objective of building an amodal perception system is not new and has long been studied separately or jointly in the community. There is plenty of literature on visual understanding despite the presence of occlusion~\cite{winn2006layout,tighe2014scene,gao2011segmentation,senior2006appearance}, depth ordering~\cite{6410421,Zhang_2015_ICCV,uhrig2016pixel}, and object completion and inpainting~\cite{ehsani2018segan,kar2015amodal}, that together show the feasibility and practicability of building machine vision systems with such capabilities. In this work, we try to solve all of these problems in a single amodal segmentation framework. 

Amodal segmentation is not an easy task, as it needs a model to understand not only semantic concepts but also the relative depth order between them. Existing researches~\cite{li2016amodal,zhu2017semantic,follmann2019learning} have tended to directly use the formulation of traditional semantic segmentation to deal with amodal segmentation, where amodal masks, as well as visible/invisible masks, are regressed for amodal object proposals. Though depth order is retrievable by analyzing the amodal mask and the visible/invisible mask, it is not explicitly supervised in the training process; thus it is not clear whether the model has learned such information. In addition, even though predicting a segmentation mask for the invisible parts of an object is more difficult than for that for the visible parts, existing methods directly regress the full amodal mask without considering such differences in difficulties. In contrast, we believe that each part of an object with different levels of occlusion should be treated separately.

\begin{figure*}[t]
	\centering
	\includegraphics[width=0.94\linewidth]{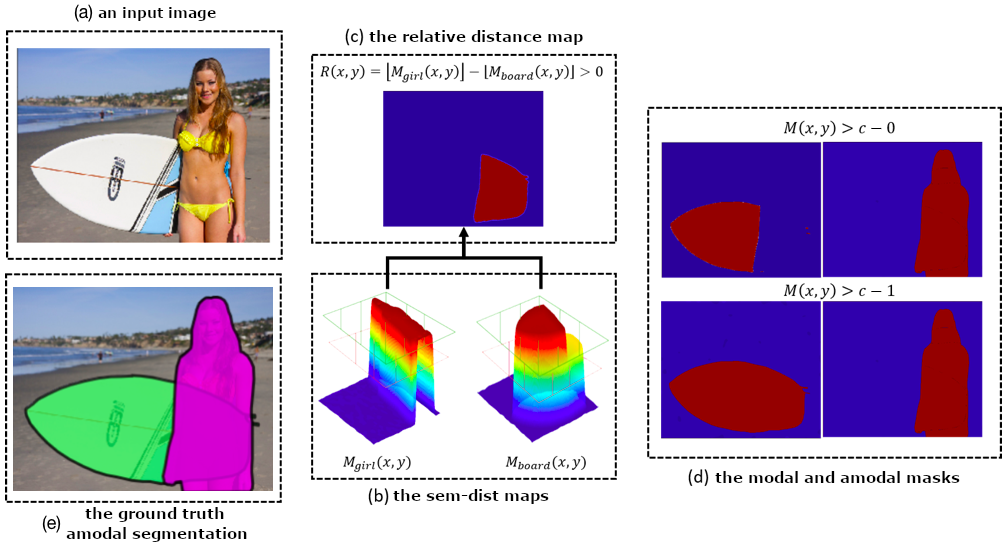}
	\caption{Illustrations of: a) a example image of a girl holding a surfboard; b) the sem-dist maps for the girl $M_{girl}(x,y)$ and the surfboard $M_{board}$; c) the relative distance map $R(x,y)$ of the girl and the surfboard; d) the masks of the girl and the surfboard generated by moving the reference level plan from the highest visibility level to the lowest as shown in (b); e) the ground truth amodal segmentation of the girl and the surfboard. (best viewed in color)}
	\label{fig:demo-of-sem-dist-map} 
\end{figure*}

To this end, we propose the formulation of the amodal segmentation problem as the learning of a \textit{semantics-aware distance map}{} (sem-dist map) for each object, which describes the pixel-level modal, amodal and relative depth order, as demonstrated in Fig.~\ref{fig:demo-of-sem-dist-map}\{a,b,d\}. The sem-dist map is a kind of level-set representation, which describes not only the confidence of occurrence but also the global visibility level of the corresponding object in a scene. Take two overlapping objects as an example (see Fig.~\ref{fig:demo-of-sem-dist-map}\{a,b,d\}). The values in their sem-dist maps should at least fall into three intervals: pixels that do not belong to each object's amodal segmentation in the backmost interval, pixels that belong to the visible parts of each object in the uppermost interval and pixels that belong to the invisible parts of occluded objects in the middle interval. In this case, imagine that we have a reference level plane that can move up and down along the intensity dimension of the sem-dist maps. If we move the plane from the highest level to the lowest, the whole object in front and the visible parts of the occluded objects, \ie, their modal segmentation, will first emerge out of the plane. Then as we move the plane down to the lower bound of the middle interval, we will have the amodal segmentation of both objects. 

This kind of formulation has several advantages: first, it unifies amodal segmentation and modal segmentation in a single framework, and one can easily get them by just thresholding the sem-dist map; second, since the relative visibility level is equivalent to the relative depth order, the output of our framework naturally contains pixel-wise depth information for each object; third, each part of an object is naturally layered on the sem-dist map according to their visibility level, which explicitly reflects the difference in difficulties in predicting their segmentation. In order to estimate a sem-dist map from an image, we introduce a novel convolutional neural network (CNN) architecture, which we refer to as the semantic layering network (SLN). The SLN is a proposal-based two-stage framework consisting of four modules: an encoder network (ENC), a global layering module (GLM), a region proposal network (RPN), and an instance layering module (ILM). An input image first goes into the ENC for feature extraction. Then the global layering maps, which describe the pixel-level visibility level for all objects, are inferred by the GLN. Finally, the global layering maps and the extracted image features are fed into both the RPN and the ILM to predict the instance layering maps and the final sem-dist maps. The pipeline of our amodal segmentation framework is shown in Fig.~\ref{fig:pipeline}, and the architecture of our proposed SLN is demonstrated in Fig.~\ref{fig:arch-sln}.

In sum, the main contributions of this work are as follows: first, we introduce the sem-dist map for amodal instance segmentation, which unifies all targets of amodal segmentation into a single compact representation; second, we present a novel CNN architecture to solve the amodal segmentation problem; and third, we conduct extensive experiments which demonstrate that our framework can predict amodal segmentation and depth order with state-of-the-art performance. We have released our code, pre-trained models and results\footnote{\url{https://github.com/apchenstu/SLN-Amodal.git}}.

\section{Related Work}
As previously mentioned, several topics, including occlusion handling, depth ordering, and object completion, are related to amodal segmentation. Hence we first briefly review the literature on these topics, then introduce the existing methods on amodal segmentation.
\subsection{Visual Understanding with Occlusion}
There has been plenty of researches on object detection~\cite{ren2015faster,liu2016ssd,lin2017feature}, semantic segmentation~\cite{chen2014semantic,long2015fully,chen2018deeplab}, and instance segmentation~\cite{pinheiro2015learning,dai2016instance,he2017mask}, yet most of them have merely focused on what we can see in an image, \ie, predicting bounding boxes around or pixels belonging to the visible parts of objects. As such, occlusion is often treated as noise and is overcome implicitly through learning-based methods that train on huge amounts of data. There are also other researchers trying to explicitly eliminate the effect of occlusion. For instance, \cite{winn2006layout} used conditional random fields (CRFs) to describe the possible configurations of object parts, which imposes constraints on the appearance of partially occluded objects given their visible parts. \cite{gao2011segmentation} introduced binary cells for object detection, thereby explicitly indicating the visibility of object parts, thus making detection more robust to occlusion. \cite{guo2012beyond} incorporated global scene priors and occluding object classes as cues to complete the labeling of partially occluded semantic regions. \cite{hsiao2014occlusion} proposed the modeling of occlusions by inferring the 3D interactions of objects for instance detection from an arbitrary viewpoint. And finally, \cite{Chen_2015_CVPR} used class-specific likelihood maps and inferred visible and occluded regions to obtain segmentation candidates via energy minimizing frameworks, which were then scored using a class-specific classifier.
\subsection{Depth Ordering}
Depth ordering has long been studied. Early research on object tracking~\cite{854907,1498758,1265863,4359366} incorporated shapes, boundaries, and occlusion patterns across frames as clues to predict depth order. In~\cite{6247688}, depth order was inferred using features extracted from boundaries and junctions separately, on which a Markov Random Field (MRF) model and graph optimization were applied to get globally consistent ordering. \cite{6410421} first identified occlusions between objects with T-junctions and highly convex contours, then used these occlusion cues to arrange objects according to depth order. \cite{Zhang_2015_ICCV} built a CNN-based architecture to jointly reason pixel-wise instance-level segmentation as well as depth order from multi-scale image patches and they combined predictions into the final labeling via the MRF. \cite{uhrig2016pixel} used a fully convolutional network (FCN) to jointly predict pixel-level semantic labels, depths and the directions to object centers from a single street scene image. 
\subsection{Object Completion and Inpainting}
Some of the research that was capable of handling occlusions~\cite{winn2006layout,gao2011segmentation,guo2012beyond,Chen_2015_CVPR} also applied object completion to occluded objects. Those methods, however, have been shown to work only on specific or limited object categories and relied on available shape models or depth inputs. \cite{kar2015amodal} took a step forward by utilizing a probabilistic framework to learn category-specific object size distributions and then leveraged the model to estimate the veridical size of the occluded objects in new images. Also, a recent piece of research~\cite{ehsani2018segan} made attempts to generate the appearance of the invisible part of an object through a generative adversarial network (GAN). For depth inpainting, research~\cite{inpaintWithColor} used color images to guide the inpainting of the corresponding depth maps by aligning the edges. Xue~\etal~\cite{xue2017depth}, on the other hand, used low-rank regularization to inpaint single depth images without using color images.
\subsection{Amodal Segmentation}
The concept of amodal segmentation has just emerged in the last few years~\cite{li2016amodal,zhu2017semantic,follmann2019learning}, though similar problems had actually been addressed years before in many applications including detection~\cite{gao2011segmentation,guo2012beyond,kar2015amodal}, segmentation~\cite{winn2006layout,Chen_2015_CVPR}, reconstruction~\cite{gupta2013perceptual,silberman2014contour}, and so on. Traditional approaches, however have usually relied on depth information or focused on specific object categories, while recent methods have solely used RGB images only and targeted objects of arbitrary categories. \cite{li2016amodal} presented the first method to tackle the amodal segmentation problem, where authors first generated data with amodal ground truths by randomly overlaying one object instance onto another and then using the generated data to train a network that could predict the segmentation heatmap. An amodal bounding box was then generated using their proposed Iterative Bounding Box Expansion strategy from the segmentation heatmap and the modal bounding box predicted by a general-purpose object detector. \cite{zhu2017semantic} built a large amodal segmentation dataset with human-annotated masks of amodal, visible and invisible regions of each object together with relative depth orders. Multiple baseline models were designed and trained on the proposed dataset to predict amodal masks or the depth orders. \cite{follmann2019learning} proposed a multi-task model that simultaneously predicted amodal masks, visible masks, and occlusion masks for each object instance. In addition, they also provided a new semantic amodal segmentation dataset D2S amodal and supplemented the class labels for the amodal datasets proposed in \cite{zhu2017semantic}.

\section{Semantics-aware Distance Map}
In this section, we will introduce the semantics-aware distance map (the sem-dist map) for amodal instance segmentation (the example sem-dist maps for a girl and a surfboard are shown in Fig.~\ref{fig:demo-of-sem-dist-map}b). Unlike the commonly used heatmap in instance segmentation, where each pixel measures the confidence of the occurrence of the visible part of an object ranging from zero to one, the sem-dist map describes the pixel-level \textit{visibility} of the \textit{whole} object where each pixel value ranges from zero to positive infinity. 


More concretely, we first define the visibility level $L$ of a region $\Omega=\{(x,y)\}$ belonging to an object to be an integer $l$ if and only if the region will become visible after we `remove' at least $l$ objects from the scene. Particularly, if a region does not belong to the object at any visibility level, we define its visibility level to be zero. Then, the sem-dist map $M$ for an object instance can be defined as:
\begin{align}
& M(x, y) = C(x,y) - L(x, y) \label{eq:def_of_sdmap}
\end{align}
where $L(x, y) \in \mathbb{N}^0$ and $C(x, y) \in [0, 1)$ stand for the visibility level and the confidence of occurrence of the object's amodal segmentation at $(x, y)$ on the image, respectively. The intuition behind this definition is that an amodal segmentation framework should have higher confidence in predicting the occurrence of an object part when it is visible in an image, and have lower confidence when it is occluded by more other objects, \ie, when the visibility level of the object part is higher. The sem-dist map emphasizes such varying difficulties in predicting segmentation for an object by explicitly adding bias to the heatmap of an object part according to its visibility level. 

The sem-dist map contains rich object information. First, it contains both modal and amodal segmentations of an object. To see that, if one only considers the region in the sem-dist map where the visibility level equals zero, then $M$ is reduced to the modal heatmap of the object $M_{modal}$. In addition, the amodal heatmap $M_{amodal}$ can also be easily calculated by taking the fractional part of each pixel in $M$, \ie
\begin{align}
& M_{modal}(x, y)  = \left\{\begin{aligned}
& M(x, y),& & M(x, y) \in [0, 1)\\
& 0  ,& & \text{otherwise}
\end{aligned}
\right.\\
& M_{amodal}(x, y) = M(x, y) - \lfloor M(x, y) \rfloor
\end{align}
where $\lfloor \cdot \rfloor$ stands for the floor function.

On the other hand, for the same region in the sem-dist maps of two objects where one occludes another, the one with a lower visibility level indicates that the corresponding part of the object is closer to the camera than another. Thus the pixel-level depth order of two objects can also be retrieved from the sem-dist map. Considering the sem-dist maps of two mutually intersecting objects $M_A$ and $M_B$, if we take the difference of the integer part of the two corresponding sem-dist maps, we will have another map which we denote as $R_{AB}$, \ie
\begin{align}
R_{AB}(x, y) = \left\{\begin{aligned}
& \lfloor M_A(x, y) \rfloor - \lfloor M_B(x, y) \rfloor, & & (x, y) \in \Omega_{AB}\\
& 0                                                    , & & \text{otherwise}
\end{aligned}\right.\label{eq:depth_order}
\end{align}
where $\Omega_{AB}$ stands for the region within which object $A$ and $B$ are mutually intersecting, \ie 
\begin{align}
\begin{aligned}
\Omega_{AB} = \{(x, y)|& \left( M_A(x, y) - \lfloor M_A(x, y) \rfloor \right) \cdot \\ 
& \left( M_B(x, y) - \lfloor M_B(x, y) \rfloor \right) > c^2\}
\end{aligned}
\end{align}
where $c$ is the confidence threshold. Note that we use $c^2$ here because that $\left( M_A(x, y) - \lfloor M_A(x, y) \rfloor \right) \cdot \left( M_B(x, y) - \lfloor M_B(x, y) \rfloor \right)$ is homogeneous to the square of the confidence $C(x, y)$, according to Eq.~\ref{eq:def_of_sdmap}.

\begin{figure}[h]
	\centering
	\includegraphics[width=0.8\linewidth,height=0.5\linewidth]{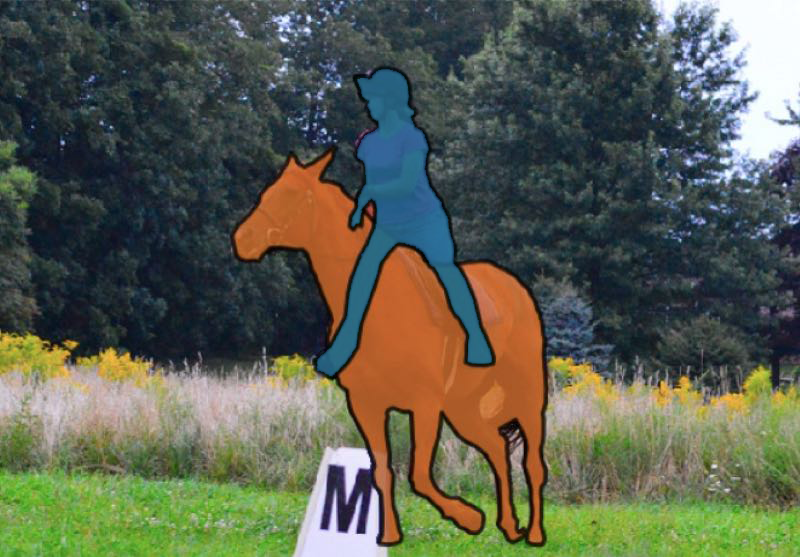}
	\caption{A figure on horseback. The pairwise depth order between the rider and the horse is unclear when two legs of the rider have different depth orders with respect to the horse. (best viewed in color)}
	\label{fig:figure-on-horseback} 
\end{figure}

We can easily get the pixel-wise depth order between $A$ and $B$ by examining the sign of each pixel in $R_{AB}$ as shown in Fig.~\ref{fig:demo-of-sem-dist-map}\{b,c\}: if $R_{AB}(x, y) > 0$ then $A$ is closer to the camera than $B$ and if $R_{AB}(x, y) < 0$ then $B$ is closer to the camera than $A$ at $(x, y)$. Taking the $R(x,y)$ in Fig.~\ref{fig:demo-of-sem-dist-map}c as an example, we can conclude that the girl is in front of the surfboard where they are occluded because $R(x,y) > 0$ in that region. Note that the depth order between $A$ and $B$ is undefined where $R_{AB}(x, y) = 0$. In addition, one should also note that the depth order is defined on a regional basis in this work, rather than on an object basis as that in the amodal dataset~\cite{zhu2017semantic}. The reason is that each part of an object can have a different depth order with respect to other objects. Considering a figure on horseback, if a picture is taken from the side of the horse, then the two legs of the rider should have different depth orders with respect to the horse (see Fig.~\ref{fig:figure-on-horseback}). Thus the depth order of an object is not well-defined, and extra criteria are needed in order to get the depth order of the full object.

\begin{figure*}[t]
	\centering
	\includegraphics[width=0.9\linewidth,height=0.35\linewidth]{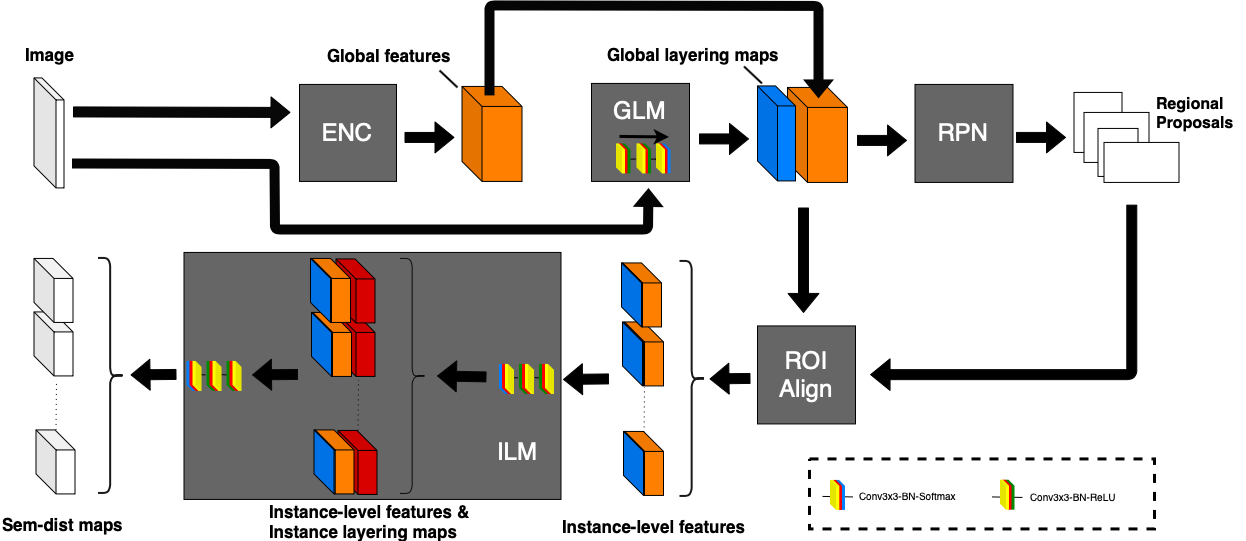}
	\caption{The architecture of semantics layering network. The input image first goes into the ENC for feature extraction. At the same time, GLM also takes the input image to predict the global layering maps. After that, both the feature maps from ENC and the global layer maps are fed into RPN to generate amodal bounding boxes proposals, which are used by ROI-Align layer \cite{he2017mask} to extract both features and global layer maps for each instance. Finally, the outputs from ROI-Align layer are collected by the ILM to estimate the final instance-level sem-dist maps. (best viewed in color)}
	\label{fig:arch-sln} 
\end{figure*}

From the discussions above, we can see that in addition to semantic information, the sem-dist map is also capable of describing the pixel-wise depth order between two objects. Given such properties of this kind of representation, we call it semantics-aware distance map (sem-dist map). Though we use the term distance to emphasize the fact that we can get the relative distance from the camera to mutually intersecting objects with a sem-dist map, one should note that the absolute distance cannot be retrieved from it. In addition, if an object has self-occluded parts on its amodal segmentation, relative distances will be determined by the occluding parts that are closer to the camera, according to our definition.

The sem-dist map is a compact representation, where the modal and amodal segmentation, occlusion and depth order of objects are well-described. Its completeness and compactness make it a desirable target for amodal perception, yet also make it difficult to train with existing methods. Therefore we introduce a carefully designed convolutional neural network (CNN) architecture to infer the sem-dist map from an image. 

\section{Semantics Layering Network}

Here we introduce the semantics layering network (SLN), a two-stage CNN-based architecture which infers instance-level sem-dist maps from a single RGB image.
\subsection{Overview}
SLN is a member of proposal-based two-stage architectures like Mask R-CNN~\cite{he2017mask}, and it is composed of four modules: an encoder network (ENC), a region proposal network (RPN), a global layering module (GLM) and an instance layering module (ILM), as shown in Fig.~\ref{fig:arch-sln}. An input image first goes into the ENC for feature extraction. At the same time, GLM also takes the input images to predict a global layering maps. After that, both the feature maps from ENC and the global layer maps are fed into RPN to generate amodal bounding box proposals, which are used by an ROI-Align layer~\cite{he2017mask} to extract both features and global layer maps for each instance. Finally, the outputs from the ROI-Align layer are collected by the ILM to estimate final instance-level sem-dist maps. Since the architecture of the RPN is directly borrowed from the Faster R-CNN~\cite{ren2015faster}, we only detail the remaining three modules in the following subsections.

\subsection{Encoder Network}
Just like most CNN-based architectures, we utilize a general-purpose encoder network to extract features from input images before task-specific modules. The ENC can be an arbitrary network which has both a large enough receptive field to capture global information and a powerful enough capability to extract low-level and high-level features. Since the requirements of the encoder for our task meet that for semantic segmentation, we choose a dilated ResNet-50~\cite{he2016deep} as our ENC architecture. In addition, we also collect features of size $256 \time 256$, $128 \times 128$, $64 \times 64$ and $32 \times 32$ from four stages of ResNet, rescale them to the same height and width, then concatenate them all together to provide the final feature maps used by the following task-specific modules.

\begin{table*}
	\centering\footnotesize\renewcommand\arraystretch{1.5}
	\addtolength{\tabcolsep}{-1.1mm}\hspace{-3mm}
	\begin{tabular}{@{\hskip 0mm}l|c|cc|ccc|c|cc|ccc|c|cc|ccc@{\hskip 0mm}}
		\multirow{2}{*}{}
		& \multicolumn{6}{c|}{all regions}
		& \multicolumn{6}{c|}{things only}
		& \multicolumn{6}{c}{stuff only}\\
		& AP & AR\tss{10} & AR\tss{100} & AR\tss{N} & AR\tss{P} & AR\tss{H}
		& AP & AR\tss{10} & AR\tss{100} & AR\tss{N} & AR\tss{P} & AR\tss{H}
		& AP & AR\tss{10} & AR\tss{100} & AR\tss{N} & AR\tss{P} & AR\tss{H}\\
		\shline
		AmodalMask~\cite{zhu2017semantic}
		& 5.74 & 13.5 & 29.23 & 34.4 & 31.0 & 21.3 
		& 6.12 & 16.5 & 33.1 & 36.2 & 37.0 & 23.6
		& 0.78 & \textbf{5.4} & 18.1 & 22.3 & 16.1 & 18.0\\
		\hline
		ARCNN~\cite{follmann2019learning}
		& 4.1 & 10.2 & 21.3 & 27.2 & 22.0 & 13.3
		& 4.4 & 12.0 & 23.9 & 28.3 & 34.7 & 15.2 
		& 0.3 & 4.8& 13.8 & 19.8 & 15.1 & 10.1\\
		\hline
		ARCNN with visible mask
		& 6.6 & 15.3 & 32.4 & 42.5 & 34.8 & 17.1
		& 7.8 & 19.5 & 37.6 & 45.5 & 40.8 & 19.9
		& 0.5 & 3.3 & 17.1 & 22.9 & 19.9 & 12.5\\
		\shline
		SLN (ours)
		& \textbf{8.4} & \textbf{16.6} & \textbf{36.5} & \textbf{44.8} & \textbf{40.1} & \textbf{22.5}
		& \textbf{9.6} & \textbf{20.5} & \textbf{40.5} & \textbf{47.2} & \textbf{43.6} & \textbf{24.9}
		& \textbf{0.8} & 5.3 & \textbf{25.0} & \textbf{28.8}& \textbf{31.3} & \textbf{18.6}\\
		\shline
		
	\end{tabular}
	\Caption{Amodal segmentation results on the COCOA validation set for our method and two state-of-the-art methods under no, partial, and heavy occlusion (AR\tss{N}, AR\tss{P}, AR\tss{H}) and for different object types (things and stuff).}
	\label{tab:eval-amodal-cocoa}
\end{table*}

\subsection{Global Layering Module}
In a proposal-based two-stage architecture, task-specific heads take cropped features after an ROI-Pooling~\cite{ren2015faster} or an ROI-Align~\cite{he2017mask} layer to make instance-level inferences. Though such a scheme is suitable for object detection or modal segmentation,\etc, where the cropped features basically contain all the required information, it is not suitable in our case. The reason is that the sem-dist map describes the \textit{global} visibility level for an object, yet the instance-level feature maps mostly contain information on single objects. Therefore, in addition to using an encoder with a larger receptive field, we introduce a global layering module (GLM) to explicitly encode the global visibility level in a global layering map.

The global layering map $M_G$ is a multi-layer heatmap, of which each layer describes the pixel-level confidence of the occurrence of an object at the corresponding visibility level determined by the index of the layer. In other words, in $M_G$, the amodal heatmap of all objects are `placed' in different layers according to the visibility level. $M_G$ will retain the information of the global visibility level for all of the objects after passing through the ROI-Align layer because such information is encoded in the channel dimension. In addition, $M_G$ can also be viewed as a pixel-level amodal proposal, which could be used to enhance the ability of RPN to propose amodal bounding boxes. Therefore $M_G$ is a desirable intermediate target for our task.

As for the architecture of GLM, we adopt a simple \textit{conv-bn-relu-conv-bn-relu-conv-softmax} block, which stands for three convolution layers followed by Batch Normalization and ReLU activation, although the last convolution layer uses a softmax activation function. The kernel size and stride size is $3 \times 3$ and $1$, respectively, for all convolution layers.

\subsection{Instance Layering Module}
The ILM aims to infer the sem-dist map for each object instance from the feature maps after the ROI-Align layer. Though the sem-dist map can be regressed directly, we decide to first predict the instance layering map as an intermediate representation, and then estimate the final sem-dist map from it. The definition of the instance layering map is similar to the global instance map, except that the former only predicts the layered amodal heatmap for a single object instance rather than for all objects. We used the same architecture as GLM to regress the intermediate instance layering map. After that, the sem-dist map is first derived from the instance layering map with Eq.~\ref{eq:def_of_sdmap}, then we use a \textit{conv-bn-relu-conv-bn-relu-conv} block which takes the derived sem-dist map and features from the ROI-Align layer to predict a refined sem-dist map. The kernel size and stride size is $3 \times 3$ and $1$ respectively, for all convolution layers.

\subsection{Loss Function}
There are three intermediate targets, \ie, the region proposals, the global layering map as well as the instance layering map, and one final target, \ie, the sem-dist map, that need to be supervised. For the region proposals, we use the smooth L1 loss and denote this loss term as $\ell_{R}$. For the global layering map, the instance layering map and the sem-dist map, we use the binary cross entropy loss and denote the three loss terms as $\ell_{G}$, $\ell_{I}$ and $\ell_{M}$ respectively. All of the targets can be jointly optimized by minimizing the overall loss $\ell$, which is the weighted sum of all of the loss terms, \ie
\begin{align}
\ell = \lambda_{R}\ell_{R} + \lambda_{G}\ell_{G} + \lambda_{I}\ell_{I} + \lambda_{M}\ell_{M} \label{eq:loss}
\end{align}
where $\lambda_{\{R,G,I,M\}}$ stands for the weight of the corresponding loss term.

\begin{figure*}[t]
	\centering
	\includegraphics[width=1.0\linewidth,height=0.53\linewidth]{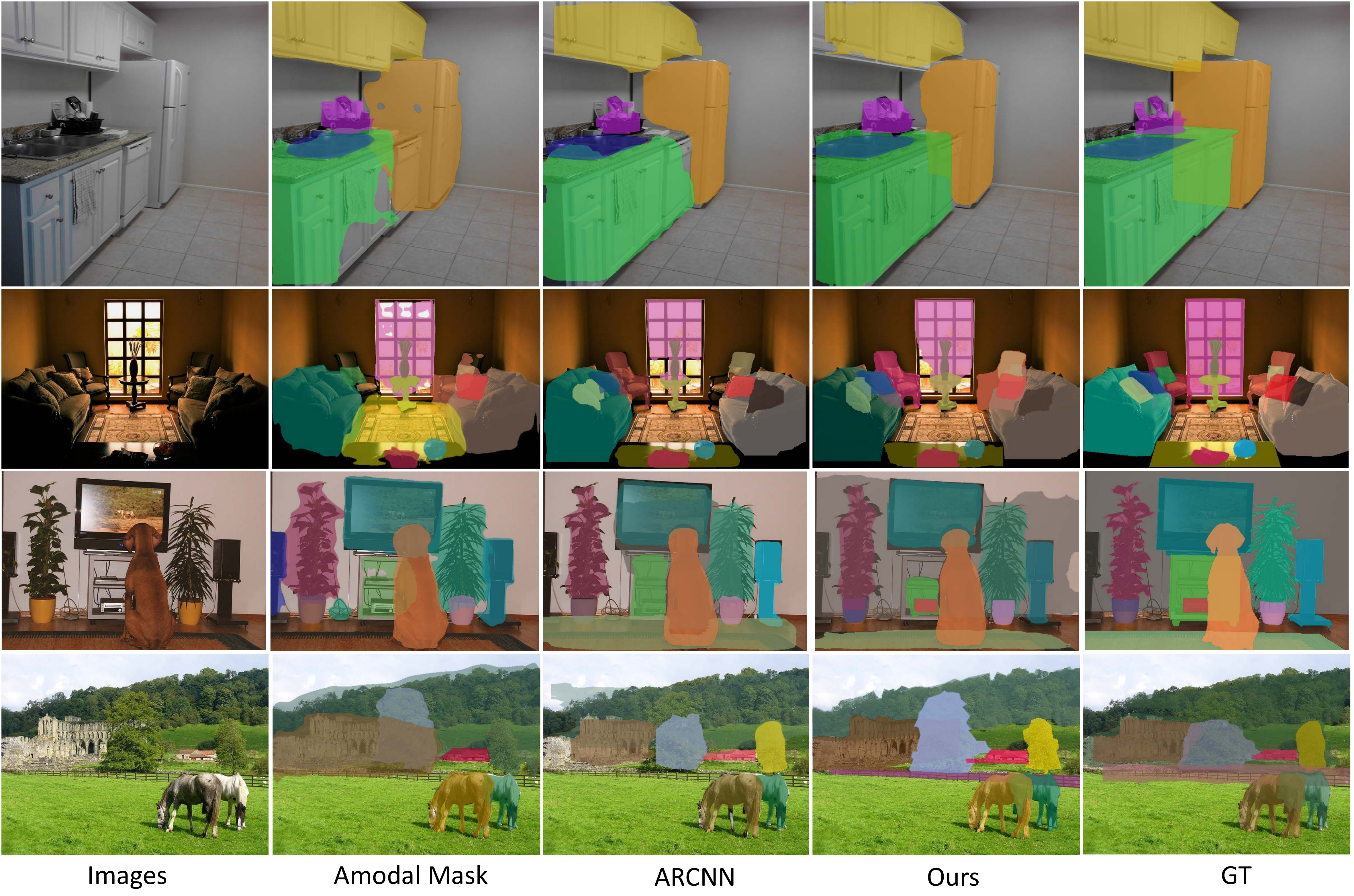}
	\caption{Some amodal segmentation results on the COCOA dataset. The columns from the left to the right are input images, results for AmodalMask~\cite{deng2017amodal}, ARCNN~\cite{follmann2019learning}, and our SLN, as well as the ground truth amodal segmentation, respectively (best viewed in color).}
	\label{fig:seg-of-cocoa}
\end{figure*}

\subsection{Implementation Details}
We implement SLN with the PyTorch \footnote{\url{https://pytorch.org/}} framework and trained it on a single Geforce 1080 Ti GPU. The ENC is initialized with parameters pretrained for semantic instance segmentation task on the COCO2014 dataset~\cite{lin2014microsoft}, and other modules are initialized with values according to the strategy described in \cite{He2015Delving}. All the targets are optimized using Stochastic Gradient Descent (SGD), with $lr=0.2, weight\_decay=5 \times 10^{-4}, momentum=0.9$, except for all normalization layers, of which $weight\_decay$ is set to zero. Although all the modules in our network could be jointly optimized with Eq.~\ref{eq:loss}, in practice we optimize GLM, RPN and ILM one after another and fix the parameter of the formerly trained module before training the later one for faster convergence, then we fine-tune all of the modules by jointly minimizing Eq.~\ref{eq:loss} with $\lambda_{\{R,G,I,M\}}=1$.

	\section{Experiments}
We conduct several experiments to evaluate the performance of our framework on the COCO amodal dataset~\cite{zhu2017semantic} and D2S amodal dataset~\cite{follmann2019learning}, which we refer to as COCOA and D2SA, respectively. The COCOA dataset is the first amodal dataset consisting of 5000 images, of which 2500, 1250 and 1250 images are used for training, validation and testing, respectively. The annotations of the COCOA dataset include the amodal segmentation of each object, visible/invisible regions of each amodal segmentation as well as the relative depth orders of all objects in each image. All of the objects in the COCOA dataset are also classified into two categories: `things' and `stuff', where a `thing' is an object with a canonical shape while `stuff' has a consistent visual appearance but can be of an arbitrary extent. The D2SA dataset is a recently proposed class-specific amodal dataset, which contains 2000 and 3600 original and artificially augmented amodally annotated images from the D2S dataset~\cite{follmann2018mvtec} for training and validation. The D2SA dataset has class labels for each object and all kinds of annotations in the COCOA dataset except for depth order. 

\subsection{Evaluation Metrics}
We evaluate the performance of our framework for amodal segmentation as well as depth order prediction, and the performance of amodal segmentation is reported for both the COCOA and D2SA datasets while the performance of depth order predictions is only reported for the COCOA dataset, due to a lacking of depth order ground truth data in the D2SA dataset.  

\textbf{Metrics for amodal segmentation:} We use the mean average precision (AP) and mean average recall (AR) as our metrics as done in previous research. The AP is calculated by averaging the precision of mask prediction over ten equally spaced intersection-over-union (IOU) thresholds from 0.5 to 0.95, as is common practice. The AR is computed by averaging the segmentation recall over the same set of thresholds. The AR for both 10 and 100 segments per image is reported, which we denote as AR\tss{10} and AR\tss{100}, respectively. For the COCOA dataset, in addition to the overall AP and AR, we also report the AR for 100 segmentations of things and stuff per image separately as well as for three occlusion levels (none, partial or heavy occlusion), which we denote as AR\tss{N}, AR\tss{R} and AR\tss{H}, respectively. For the D2SA dataset, the AP, AR\tss{10} and AR\tss{100} for all objects and occluded objects are reported.

\textbf{Metrics for pairwise depth order: } For the COCOA dataset, we evaluate our depth order predictions. We report the accuracy in predicting which of two overlapping objects is in front of the other. There are 36k/23k overlapping objects in the train/val set.

\subsection{Implementation Details}

\textbf{Amodal segmentation:} We compare SLN with SOTA segmentation models, including AmodalMask in~\cite{zhu2017semantic} and ARCNN in~\cite{follmann2019learning}. In all of our experiments, we use the released evaluation code\footnote{\url{https://github.com/Wakeupbuddy/amodalAPI}} used in \cite{zhu2017semantic} to compute the AP and the AR. For AmodalMask, we directly use the released segmentation results to conduct our evaluation. For ARCNN, because neither the model, the segmentation results nor the evaluation code have been released at the time of writing this paper, we re-implement ARCNN according to the descriptions in \cite{follmann2019learning} and evaluate the segmentation result using the same evaluation code in \cite{zhu2017semantic}. Although ARCNN is actually the standard Mask R-CNN~\cite{he2017mask} with a ResNet-101~\cite{he2016deep} backbone trained on the COCOA/D2SA dataset, we can not reproduce similar AP and AR results as reported in \cite{follmann2019learning}. Therefore, we report only the results of our re-implemented ARCNN and refer readers to their original paper~\cite{follmann2019learning} for their reported results. In addition, we also add another version of ARCNN where both amodal and visible masks rather than amodal masks only are regressed.

\textbf{Depth ordering:} We compare SLN with multiple baseline methods proposed in \cite{zhu2017semantic}, including SharpMask~\cite{pinheiro2016learning} and ExpandMask~\cite{zhu2017semantic}, for depth ordering. In addition, the pairwise depth order predicted with the ground truth amodal mask and the sem-dist map are also evaluated. Pairwise depth orders are extracted from amodal masks or sem-dist maps in different ways. For the amodal mask, the best matching mask is first selected for each object (with largest IOU). Then the selected masks for each pair of overlapping objects are fed into the OrderNet~\cite{zhu2017semantic}, which is a pre-trained Resnet-50 model slightly modified for varying number of input channels, to predict the pairwise depth order. The OrderNet is trained and tested separately for each set of masks. For the sem-dist map, we first extract the amodal masks and find the best matching sem-dist maps for each object. Then we use both $R(x, y)$ in Eq.~\ref{eq:depth_order} and OrderNet to predict the pairwise depth orders from their sem-dist map pairs. Recall that the depth order derived from $R(x, y)$ is defined on a regional basis rather than an object basis as in the COCOA dataset. After we get a region-level relative depth order with $R(x, y)$, an object-level relative depth order is chosen to be consistent with the relative depth order of the largest region. 

\begin{figure}[t!]
	\centering
	\includegraphics[width=0.8\linewidth,height=1.0\linewidth]{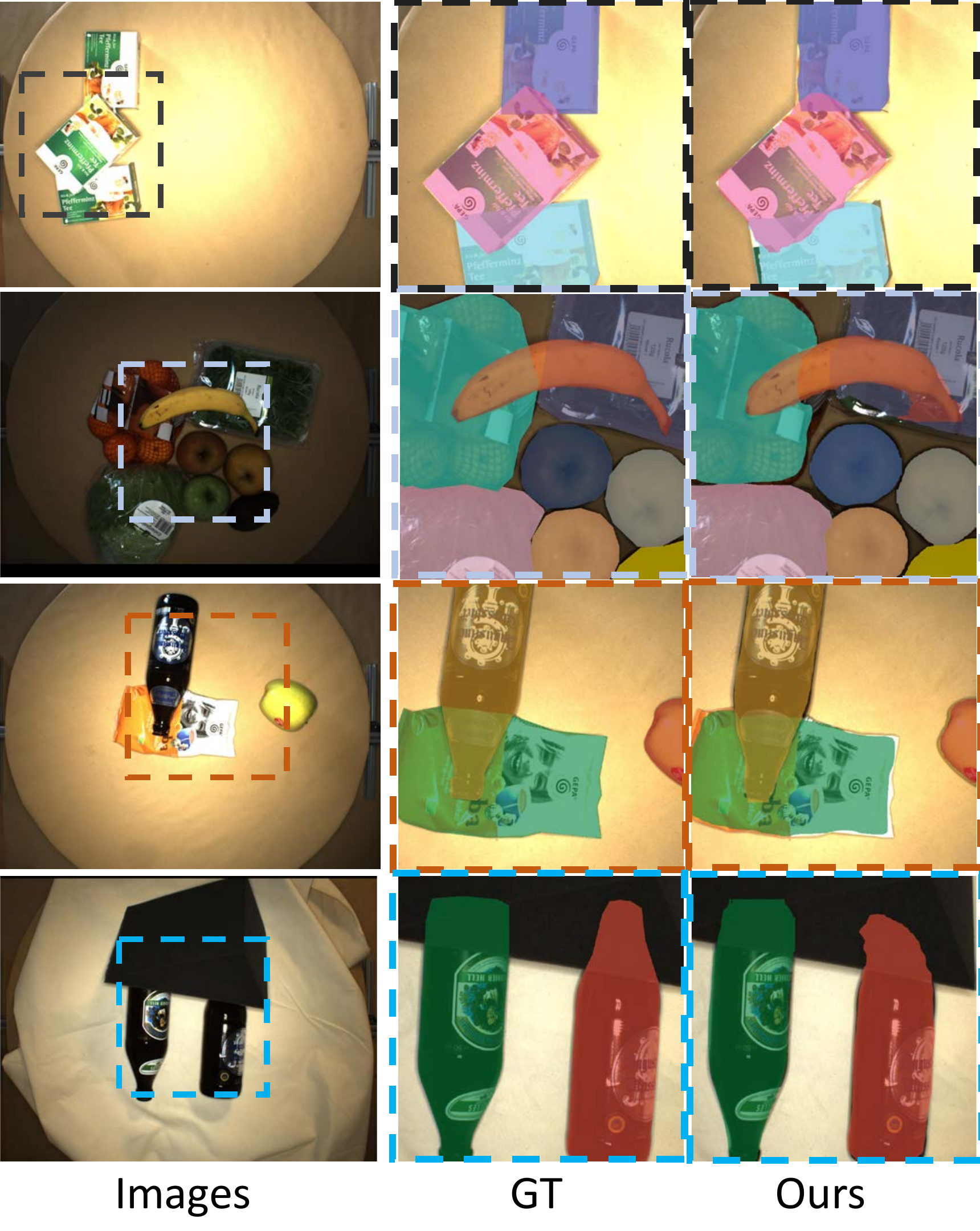}
	\caption{Some amodal segmentation results for SLN on D2SA dataset.}
	\label{fig:seg-of-d2sa}
\end{figure}

\subsection{Results}
\textbf{COCOA:} The amodal segmentation results for AmodalMask~\cite{zhu2017semantic}, ARCNN~\cite{follmann2019learning}, and our proposed SLN are listed in Table~\ref{tab:eval-amodal-cocoa}. In this experiment, our framework generally shows to achieve better recall than all of the existing methods. Interestingly, we found that the AP for all methods that rely on bounding box-based amodal proposals is lower than the one for methods that rely on mask-based amodal proposals. Besides, the former set of methods seems to work better for zero occluded and partially occluded things, while the later works better for heavily occluded stuff. We suspect that pixel-level proposals could be more easy to be generalized to the amodal case, especially for stuff, which have consistent appearance but can be of arbitrary extent. The fact that our framework achieves superior performance to ARCNN~\cite{follmann2019learning} demonstrates the effectiveness of the sem-dist map to guide the network to perceive amodal concepts. We also show some typical qualitative results for AmodalMask, ARCNN and our method on COCOA dataset in Fig.~\ref{fig:seg-of-cocoa}. From the first row, one can observe that our SLN segments the refrigerator and the cupboard better and the washbowl is also well recognized. For the second raw, the SLN not only accurately segments the sofa, but also identified the bolster on it. And for the forth row, our SLN is the only method that correctly completes the occluded part of the horse.


As for depth ordering, the comparison of accuracy between SLN with multiple baselines are listed in Table~\ref{tab:depth-order-result}. Because sem-dist maps contain rich depth information, the OrderNet predicts depth order better with the predicted sem-dist maps than masks. Besides, $R(x, y)$ is also able to give us reasonable depth order predictions. 

\textbf{D2SA:} The results on D2SA dataset are listed in Table~\ref{tab:eval-amodal-d2sa}. We can observe that our framework achieves better overall results than ARCNN for both AP and AR. Examples of amodal mask predictions for ARCNN and our method are shown in Fig.~\ref{fig:seg-of-d2sa}. As one can see, our method is able to retrieve the occluded parts of objects and to predict reasonable amodal segmentation. Interestingly, SLN works better for heavy occlusions on D2SA dataset than on COCOA dataset. This is because there are only things in the D2SA dataset and the experiments on both datasets indicate that SLN (and other SOTA methods that rely on bounding box proposals) works pretty well for "things" rather than "stuff".

\begin{table}[t]
	\centering\footnotesize\renewcommand\arraystretch{1.5}
	\addtolength{\tabcolsep}{-1.5mm}
	\begin{tabular}{@{\hskip 0mm}l|c|c|c|c|c|c@{\hskip 0mm}}
		& Sharp                             & Expand                       & Amodal                        & SLN & Ground  & Ground \\
		& Mask~\cite{pinheiro2016learning}  & Mask~\cite{zhu2017semantic}   & Mask~\cite{zhu2017semantic}    & Map & TruthM  & TruthS \\
		\shline
		OrderNet~\cite{zhu2017semantic} 
		& .786 & .785 & .791 & \textbf{.854} & .817 & \textbf{.872} \\
		\hline
		$R(x,y)$
		&  -   &  -   &  -   & .764 &  -   & 1.00 \\
		\shline
	\end{tabular}
	\Caption{Accuracy of pairwise depth order predicted by multiple baselines proposed in~\cite{zhu2017semantic} and our method. Note that the inputs of the OrderNet are masks for all mask-based methods and sem-dist maps for our method. The GroundTruthM stands for the ground truth masks and the GroundTruthS stands for the ground truth sem-dist maps.}
	\label{tab:depth-order-result}
\end{table}

\begin{table}[t]
	\centering\footnotesize\renewcommand\arraystretch{1.5}
	\addtolength{\tabcolsep}{-1.1mm}\hspace{-3mm}
	\begin{tabular}{@{\hskip 0mm}l|c|cc|c|cc|c|cc@{\hskip 0mm}}
		\multirow{2}{*}{}
		& \multicolumn{3}{c|}{all}
		& \multicolumn{3}{c|}{partial occlusion}
		& \multicolumn{3}{c}{heavy occlusion}\\
		& AP & AR\tss{10} & AR\tss{100}
		& AP & AR\tss{10} & AR\tss{100}
		& AP & AR\tss{10} & AR\tss{100}\\
		\hline
		ARCNN\cite{follmann2019learning}
		& 23.4 & 47.3 & 73.1
		& 6.2 & 38.0 & 72.6
		& \textbf{0.6} & 17.8 & 45.0\\
		\hline
		SLN (ours)
		& \textbf{25.3} & \textbf{41.3} & \textbf{78.6}
		& \textbf{6.5} & \textbf{42.4} & \textbf{77.5}
		& 0.6 & \textbf{22.7} & \textbf{57.2}\\
		\shline
		
	\end{tabular}
	\Caption{The quantitative results for ARCNN and our method on D2SA validation set for all/occluded objects.}
	\label{tab:eval-amodal-d2sa}
\end{table}

\section{Conclusion}
In this work, we proposed to tackle the amodal segmentation problem by learning a semantics-aware distanc e map (sem-dist map) for each object in an image. Compared with the commonly used mask representation, the semantics-aware distance map describes pixel-level visibility level of an object, from which the modal, amodal segmentation and relative depth order of the object can be derived elegantly. In order to estimate the sem-dist map, we introduced the semantics layering network (SLN), in which sem-dist maps for all objects are inferred layer by layer, from global-level to instance-level, from an image. Extensive experiments on COCOA and D2SA datasets have demonstrated that our framework can predict amodal segmentation and pairwise depth order with state-of-the-art performance. 

In our experiments, we observed that the performance bottleneck of SLN is the low quality of object proposals from the bounding box-based proposal modules (RPN)
It seems that RPN is not good at making large area amodal proposals. To tackle this problem, one may try to use pixel-level proposal modules or proposal-free frameworks to further boost the performance of SLN for amodal segmentation. 

\bibliographystyle{ACM-Reference-Format}
\bibliography{bibfile}

\end{document}